%% file: Causal-UAV.tex
\documentclass[letterpaper, 10 pt, conference]{ieeeconf}  

\IEEEoverridecommandlockouts                              

\usepackage{graphicx}
\usepackage{bm}
\usepackage[linesnumbered,lined,algoruled]{algorithm2e}
\usepackage{booktabs}
\usepackage{multirow}
\usepackage{cite}
\usepackage{fontawesome}
\usepackage{amsmath} 
\usepackage{amssymb}  
\usepackage{float}
\UseRawInputEncoding

\title{\LARGE \bf
Robust Policy Learning for Multi-UAV Collision Avoidance \\ with Causal Feature Selection
}

\author{Jiafan Zhuang$^{1}$, Gaofei Han$^{1}$, Zihao Xia$^{1}$, Boxi Wang$^{1}$, Wenji Li$^{1}$, Dongliang Wang$^{1}$ \\ Zhifeng Hao$^{1}$, Ruichu Cai$^{2}$ and Zhun Fan\textsuperscript{3,4, \faEnvelopeO}
\thanks{\textsuperscript{\faEnvelopeO} Corresponding author}
\thanks{*This work is supported in part by the National Science and Technology Major Project (grant numbers 2021ZD0111501, 2021ZD0111502), the National Natural Science Foundation of China (grant numbers 62176147, 51907112, U2066212, 61961036, 62162054), Science and Technology Planning Project of Guangdong Province of China (grant numbers 2023A1515011574, 2022A1515110566, 2022A1515110660), and the STU Scientific Research Foundation for Talents (grant numbers NTF21001, NTF21052, NTF22030).}
\thanks{$^{1}$College of Engineering, Shantou University}
\thanks{$^{2}$School of Computer Science, Guangdong University of Technology}
\thanks{$^{3}$University of Electronic Science and Technology of China}
\thanks{$^{4}$International Cooperation Base of Evolutionary Intelligence and Robotics}
}

\begin{document}

\newcommand{\etal}{\textit{et al}.}
\newcommand{\ie}{\textit{i}.\textit{e}.}
\newcommand{\eg}{\textit{e}.\textit{g}.}
\newcommand{\etc}{\textit{etc}}

\maketitle
\thispagestyle{empty}
\pagestyle{empty}

\begin{abstract}

In unseen and complex outdoor environments, collision avoidance navigation for unmanned aerial vehicle (UAV) swarms presents a challenging problem. It requires UAVs to navigate through various obstacles and complex backgrounds.
Existing collision avoidance navigation methods based on deep reinforcement learning show promising performance but suffer from poor generalization abilities, resulting in performance degradation in unseen environments.
To address this issue, we investigate the cause of weak generalization ability in DRL and propose a novel causal feature selection module. 
This module can be integrated into the policy network and effectively filters out non-causal factors in representations, thereby reducing the influence of spurious correlations between non-causal factors and action predictions.
Experimental results demonstrate that our proposed method can achieve robust navigation performance and effective collision avoidance especially in scenarios with unseen backgrounds and obstacles, which significantly outperforms existing state-of-the-art algorithms.

\end{abstract}

\section{INTRODUCTION}


In recent years, unmanned aerial vehicle systems~\cite{yuan2021online,cai2021behavior,li2021noisy} have made remarkable progresses and have been applied in many fields, such as agriculture~\cite{ju2019modeling,tsouros2019review}, search and rescue~\cite{tian2020search,tomic2012toward},  the mining industry~\cite{shahmoradi2020comprehensive} and patrol inspection~\cite{kim2020fault}. 
To enable collaboration among multiple UAVs, it is important to find an optimal path to the target position while avoiding collisions, especially when the swarm size is large~\cite{huang2019collision}.
Therefore, multi-UAV collision avoidance has become a fundamental and critical task, increasingly attracting the interest of researchers.

The traditional approaches~\cite{trujillo2018cooperative,lu2018survey,hrabar2011reactive} for multi-UAV collision avoidance primarily rely on real-time simultaneous localization and mapping (SLAM)~\cite{bryson2007building}. These methods utilize sensors such as LiDAR to perceive the surrounding environment and generate trajectories through path planning~\cite{aggarwal2020path}. Additionally, prior maps~\cite{khairuddin2015review,kummerle2011large,vysotska2017improving} are often introduced to enhance the performance of SLAM systems.
However, these traditional methods usually require considerable computational resources and constrained by the lack of prior maps, making it difficult to adapt to unknown environments.

\begin{figure}[t]
    \begin{center}
        \includegraphics[width=1\linewidth]{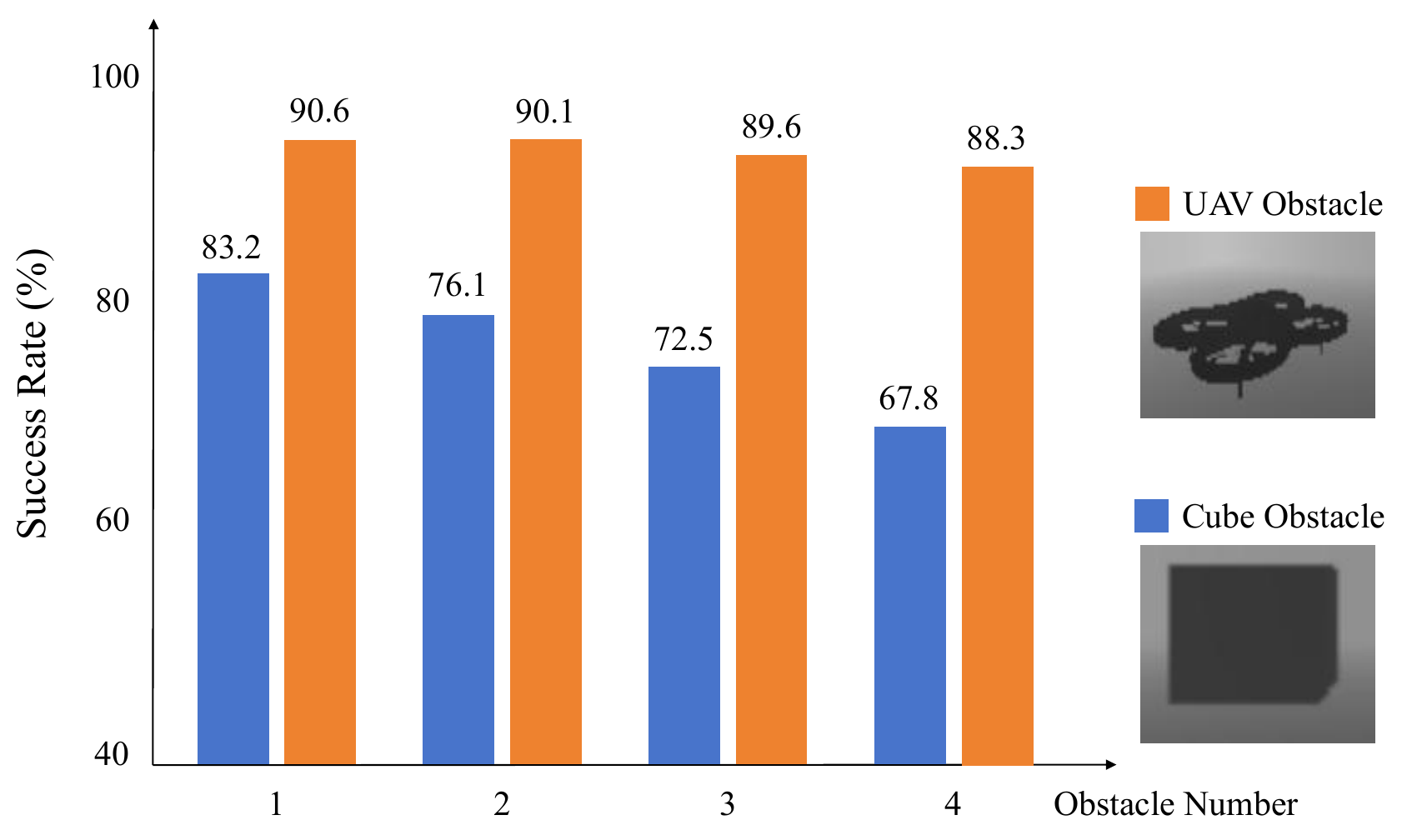}
    \end{center}
    \caption{\textbf{The influence of obstacle shape on success rate}. For the DRL model, UAV obstacles have been seen during model training while cube obstacles are unseen. During testing, the former has little influence on performance while the latter would significantly reduce the success rate of navigation.}
    \label{fig:analysis}
    \vspace{-2mm}
\end{figure}

\begin{figure}[t]
    \begin{center}
        \includegraphics[width=0.9\linewidth]{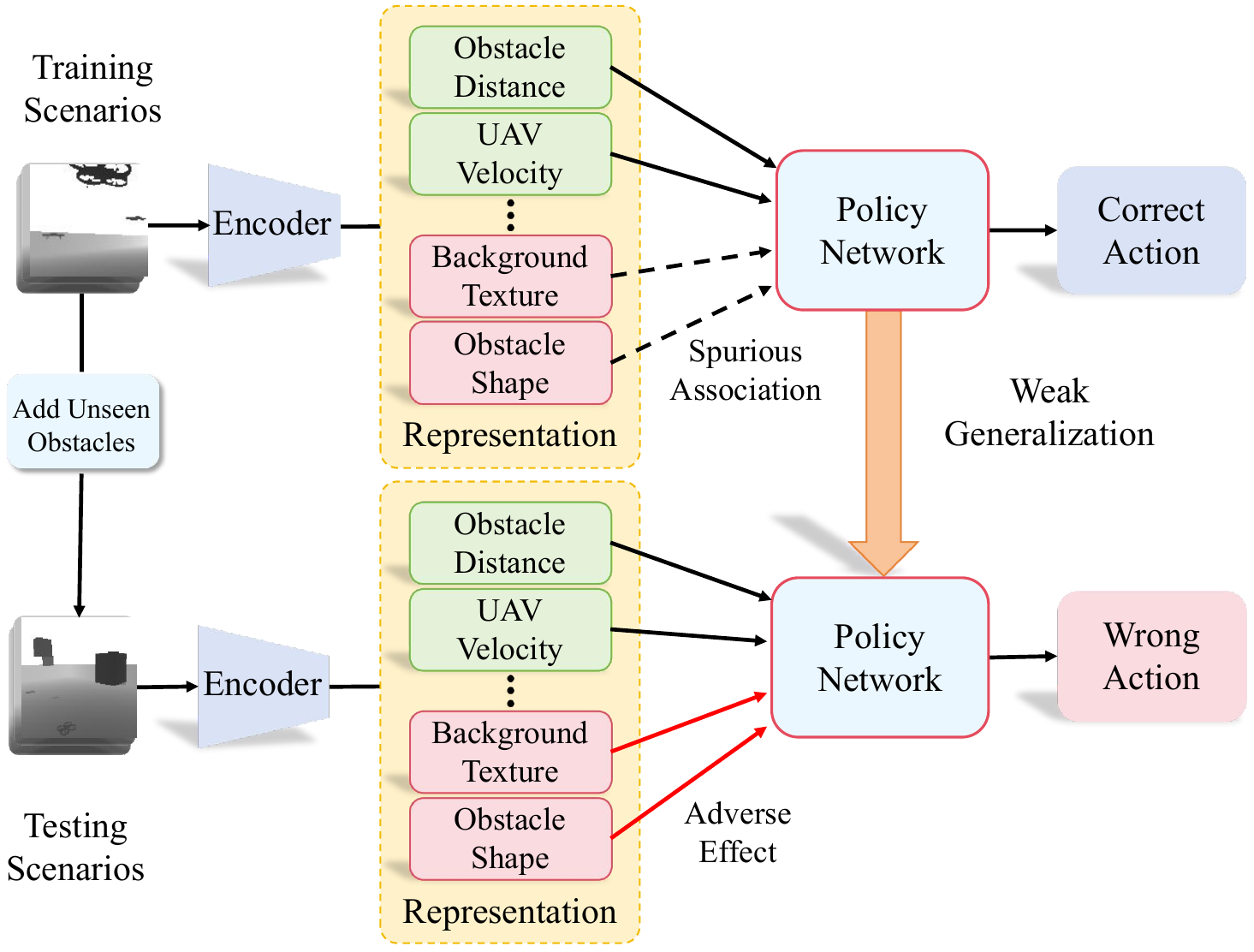}
    \end{center}
    \caption{\textbf{Illustration on the influence of non-causal representation factors}. During policy learning, non-causal factors would construct spurious correlations with action prediction. When testing scenarios are different from the training scenarios, these non-causal factors would bring adverse effect and result in wrong actions.}
    \label{fig:motivation}
    \vspace{-4mm}
\end{figure}

To address this issue, deep reinforcement learning (DRL)~\cite{arulkumaran2017deep} has been extensively studied and applied in the field of robotics. 
DRL enables end-to-end collision avoidance navigation~\cite{hu2023efficient} using only sensor information as input, eliminating the reliance on prior maps of environments.
Specifically, DRL learns visual representations from raw input data and maps these representations to an optimal flight strategy by designing appropriate policy networks and reward functions.
However, DRL is fundamentally a data-driven method that typically assumes training and testing data are sampled from an independent and identically distribution (IID) environments~\cite{hoadley1971asymptotic}.
This IID assumption can be easily violated in real-world applications, since the deployment environment has to meet the practical requirements~\cite{wu2018cooperative}, which is commonly different from the training environment.
In such a case, a generalization issue~\cite{zhou2022domain} will occur.


To investigate the generalization issue of DRL, we revisit the pioneering work of DRL-based multi-UAV collision avoidance, namely, SAC+RAE~\cite{huang2022vision}. We provide an interesting analysis, as shown in Fig.~\ref{fig:analysis}. 
SAC+RAE simulates eight UAVs for collision avoidance experiments in the training stage.
During training, each UAV treats the other UAVs as the main obstacles, allowing the model to learn a promising strategy for avoiding collisions.
In the testing stage, we introduce several UAV-like obstacles and unseen cube obstacles into the scene. Surprisingly,  the UAV-like obstacles barely interfere with the flight, while the cube obstacles significantly reduce the success rate of navigation.
This experiment demonstrates that the DRL model mistakenly constructs a relationship between the obstacle shape and the learned strategy. When encountering unseen obstacles (\eg, cube obstacles), the model fails to provide an effective strategy, significantly diminishing the generalization ability of DRL models. 


To address the generalization issue of DRL, we analyze the structure of SAC+RAE and find that the root cause of its weak generalization ability lies in the unstable and error-prone visual representations.
As illustrated in Fig.~\ref{fig:motivation}, SAC+RAE employs a regularized auto-encoder (RAE)~\cite{ghosh2019variational} to encode input depth images into compact visual representations. 
The regularized auto-encoder, trained under reconstruction supervision, tends to implicitly encode all visual information into the representations, such as obstacle distance, UAV velocity and obstacle shape.
Among these encoded pieces of information, some are useful and necessary for the collision avoidance task, such as obstacle distance and UAV velocity, which we refer to as \emph{causal factors}. In contrast, some information is specific to particular environments and irrelevant to the task, such as obstacle shape and background texture, referred to as \emph{non-causal factors}.
If we do not differentiate between these representation factors and pass them all to the policy network, spurious correlations~\cite{ward2013spurious,de2019causal} could be formed between non-causal factors and predicted actions.
Consequently, if the testing environment changes, for example by introducing some unseen obstacles, the non-causal factors could adversely affect policy predictions through the spurious correlations, leading to collisions. This is the primary cause of the weak generalization ability of DRL models to unknown scenarios.

Therefore, effectively filtering out non-causal components from visual representations is crucial for enhancing the generalization ability of DRL. To this end, causal representation learning~\cite{scholkopf2021toward,cai2019learning,li2024subspace} offers a viable solution and is a research focus in the field of artificial intelligence. It aims to identify causal representation factors by constructing underlying causal structures, effectively addressing generalization issues in out-of-distribution (OOD) scenarios~\cite{hsu2020generalized,lu2021invariant}.
Inspired by causal representation learning, we design a plug-and-play and lightweight \textbf{C}ausal \textbf{F}eature \textbf{S}election (CFS) module to be integrated into the policy network. The CFS module includes a differentiable mask that can effectively filter out non-causal feature channels. It is commonly assumed that different channels contain various visual factors~\cite{de2019causal}. To learn the optimal collision avoidance strategy, we guide the differentiable mask to maximize the action value function Q-value. Consequently, only the causal representation factors are passed on for policy learning, explicitly minimizing the impact of non-causal factors and enhancing generalization ability.


We construct several typical testing scenarios containing unseen backgrounds and obstacles to evaluate the effectiveness of our proposed method in enhancing generalization ability.
The experimental results demonstrate that our method achieves a significant improvement in the success rate of collision avoidance for UAVs over previous state-of-the-art methods, validating the superiority and effectiveness of our approach.

\section{RELATED WORK}

\subsection{DRL-based Collision Avoidance Navigation}
Research on DRL-based collision avoidance navigation is a prominent topic in the field of UAVs.
To address the scarcity of drone simulators closely resembling real environments, Cetin~\etal~\cite{cetin2019drone} introduce a drone simulation platform named Airsim built on the Unreal Engine. They utilize RGB images as observation signals and propose a discrete space-based deep deterministic policy gradient (DDPG) algorithm for drone collision avoidance navigation. However, this algorithm is trained and tested solely within the same urban environment block. 
Huang~\etal~\cite{huang2019autonomous} replace RGB images with depth images as observation states to bridge the gap between simulation and the real world. They train a value network using the deep q-network (DQN) algorithm based on a discrete action space. This method successfully transitions from a simulated environment to a real world one by training the network with a pillar matrix as the training environment and employing a new indoor space as the testing environment.
To overcome the stuttering issue caused by discrete action space in drone motion, Xue~\etal~\cite{xue2021vision} propose a vision-based collision avoidance approach using soft actor-critic (SAC) in a continuous action space. They combine SAC with a variational auto-encoder (VAE) and train the drone to perform collision avoidance tasks in a simulated environment featuring multiple wall obstacles. 

Different from existing works, we focus on examining the generalization issue of DRL-based collision avoidance navigation. We explore the relationship between visual representation and weak generalization ability, and then propose a novel causal feature selection module to address the issue of spurious correlations.

\subsection{Causal Representation Learning}
Traditional representation learning methods often rely on the  assumption of independent and identically distributed data. However, in practical applications, it is difficult to ensure that the deployment scenarios are identical to the training one. When the data deviates from the IID assumption, the performance of current machine learning algorithms tends to deteriorate significantly. Causal representation learning aims to discover causal factors by constructing the underlying causal structures, effectively addressing the generalization issue in out-of-distribution (OOD) scenarios.
Lin~\etal~\cite{lin2022causal} conduct unsupervised video anomaly detection tasks from the perspective of causal inference. They use a causal inference framework to mitigate the impact of noisy pseudo-labels on anomaly detection results.
Liu~\etal~\cite{liu2022towards}decouple the physical laws, mixed styles, and non-causal false features in pedestrian motion patterns to ensure the robustness and reusability of pedestrian motion representations. This approach addresses the problem of motion trajectory prediction.
Huang~\etal~\cite{huang2021adarl} propose an adaptive reinforcement learning algorithm that utilizes a graph model to achieve minimal state representation, incorporating the varying factors in specific domains and shared representations in common domains. This algorithm achieves robust and effective policy transfer with only a small number of samples from the target domain.
Yang~\etal~\cite{yang2021causalvae} propose to enhance the traditional VAE model to influence the generated representation through causally structured layers, but it requires the true causal variables as labels.

In this article, we are the first to apply causal representation learning within a DRL-based collision avoidance navigation framework and propose a causal feature selection module. This module can effectively address the generalization issue when UAVs are deployed in unknown scenes.


\section{APPROACH}

\begin{figure*}
    \begin{center}
    \includegraphics[width=0.73\linewidth]{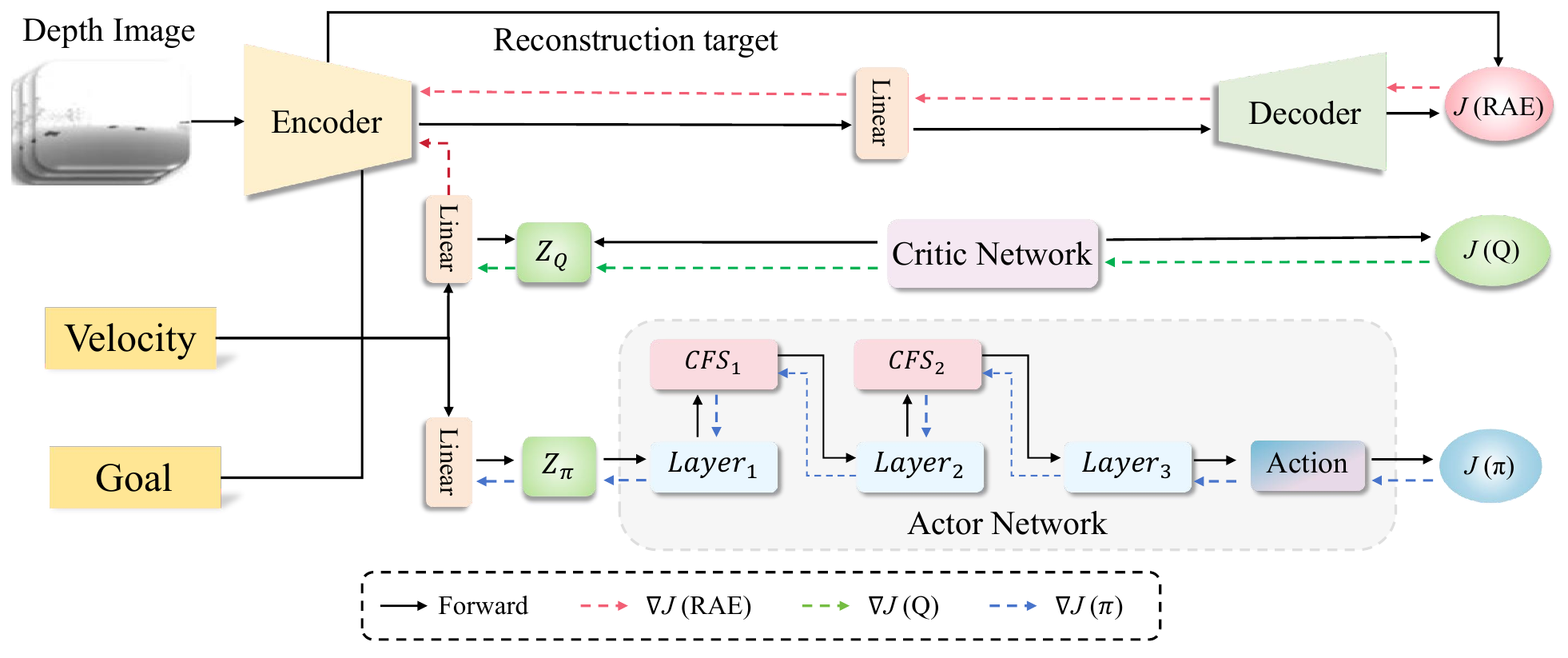}
    \end{center}
    \caption{\textbf{The architecture of our framework for Multi-UAV collision-avoidance}. The framework follows the SAC paradigm and uses an regularized auto-encoder for visual representation extraction, which takes depth images, current velocity, and relative goal position as input and outputs flight control actions.
    In the actor network, we insert our design CFS module for feature selection.}
    \label{fig:overview}
    \vspace{-4mm}
\end{figure*}

\subsection{Problem Formulation and DRL Setting}
This study aimes to empower UAVs with the capability to adapt to unseen environments by learning robust collision avoidance strategy. The UAVs can receive depth images through a front-central camera and pose information by inertial measurement unit, leading to restricted environment observation during the interaction process. Hence, the multi-UAVs collision avoidance is categorized as a partially observable markov decision process (POMDP).
The POMDP can be described as a 6-tuple $(\mathcal{S}, \mathcal{A}, \mathcal{T}, \mathcal{R}, \Omega, \mathcal{O})$, where $\mathcal{S}$ and $\mathcal{A}$ are the state and action space, respectively.
$\mathcal{T}$ is the state transition probability, which describes the probability distribution of transition from current state to next state.
$\mathcal{R}$ is the reward function.
$\Omega$ is a set of environmental observations $(o \in \Omega)$ and $\mathcal{O}$ is the observation function which describes the mapping from states to observations.

\subsubsection{Observation space}
At each timestep $t$, the observation information obtained by the agent is ${o}^{t}=[{o}_{z}^{t}$, ${o}_{g}^{t}$, ${o}_{v}^{t}]$.The ${o}_{z}^{t}$
refers to depth images taken from the camera of the UAVs, which contains distance information from a viewpoint to the surfaces of scene objects.
The ${o}_{g}^{t}$ denotes the target position information in the body coordinate system. ${o}_{v}^{t}$ denotes the velocity of the UAVs.

\subsubsection{Action space}
To enhance the diversity and controllability of UAVs, we utilize a continuous action space. At each timestep ${t}$, each UAV generates an action command with three degrees of control: $a=[{v}_{x}^{cmd}$, ${v}_{z}^{cmd}$, ${v}_{y}^{cmd}]$. The ${v}_{x}^{cmd}$ denotes forward velocity in the x-axis of the body coordinate system. Similarly, the ${v}_{z}^{cmd}$ denotes climb velocity in the z-axis of the body coordinate system. The ${v}_{y}^{cmd}$ represents steering velocity in the y-axis of the body coordinate system.

\subsubsection{Reward function}
Sparse rewards can increase the difficulty of reinforcement learning (RL) problems. To address the collision avoidance problem for UAVs, this study designs a non-sparse reward function. This function provides abundant reward signals during the training process, ensuring the convergence of RL algorithms as the agent explores the environment.
The reward function primarily comprises an obstacle avoidance reward and an arrival reward. Detailed
explanations for these rewards are provided below:
\begin{equation}
    {r}^{t}={r}^{t}_{goal}+{r}^{t}_{avoid}
\end{equation}
The reward ${r}^{t}_{goal}$ is used to guide the UAVs towards their goal. If a UAV is within a distance of less than 0.5 meters from its target position at the current timestep $t$, it will receive an arrival reward ${r}_{arrival}$.
For the $i^{th}$ UAV, if the distance between the goal ${g_{i}}$ and the current position ${p}_{i}^{t}$ is smaller than the distance between the goal and the previous position ${p}_{i}^{t-1}$, a reward ${r}^{t}_{goal}$, scaled by a rewarding weight ${\omega}_{goal}$ with the difference of the distances, will be added.
\begin{equation}
    {r}^{t}_{goal}=\left\{
        \begin{array}{l}
            {r}_{arrival} \quad \quad \quad \quad \quad \quad \quad \quad if \  \Vert {p}_{i}^{t}-{g}_{i} \Vert < 0.5 \\
            {\omega}_{goal} \cdot (\Vert {p}_{i}^{t-1}-{g_{i}} \Vert - \Vert {p}_{i}^{t}-{g}_{i} \Vert)  \ otherwise
        \end{array}
        \right.
\end{equation}
The reward ${r}^{t}_{avoid}$ is designed to guide each UAV to avoid collisions.
The UAV incurs a penalty ${r}_{collision}$ when colliding with other UAVs or obstacles in the environment. 
Alternatively, the UAV gets a penalty if the minimum distance in the depth image ${d}^{t}_{min}$ is less than the safe distance ${d}_{safe}$, which is scaled by a penalty weight ${\omega}_{avoid}$:

\begin{equation}
    {r}^{t}_{avoid}=\left\{
        \begin{array}{l}
            {r}_{collision} \quad \quad \quad \quad \quad \quad \quad if \ \ collision \ \ occurs \\
            {\omega}_{avoid} \cdot {max}({d}_{safe}-{d}^{t}_{min}, 0) \quad otherwise
        \end{array}
        \right.
\end{equation}

\subsection{Overview}
The framework of our method is illustrated in Fig.~\ref{fig:overview}. It adopts the SAC paradigm and comprises two main modules: the regularized auto-encoder and RL policy module. The regularized auto-encoder is  tasked with extracting visual representation from depth images. These visual representations, along with the current velocity and the relative goal position, are then fed into the RL policy module (\ie, actor and critic networks) for strategy learning. 
At each timestep, the RL module outputs a set of actions for collision avoidance navigation of UAVs.

Building upon the previous work~\cite{huang2022vision}, the encoder features four convolutional layers and one fully connected (FC) layer, which transforms the input depth images into a 50-dimensional latent representation. The decoder, mirroring the encoder's structure, includes one FC layer and four deconvolutional layers, reconstructing the latent representation back into the original image.

The RL policy module is divided into an actor network and a critic network. The actor network is constructed as a 3-layer multi-layer perceptron (MLP) featuring a 1024-dimensional hidden layer.
The critic network's output is generated by a 3-layer MLP with 1024 units.

To enhance policy learning, we employ the CFS module to eliminate non-causal factors from the visual representations and mitigate the effect of spurious correlations within the actor network.
The CFS module creates a binary mask that is applied to intermediate features in the actor network,
effectively facilitating feature selection.

During the training phase, the parameters of actor network are updated through the loss function $J(\pi)$, which can be expressed as follows:
\begin{equation}
    J(\pi)=\mathbb{E}_{{o}\thicksim\mathcal{B}}\left[D_{KL}(\pi(\cdot|{o})\| \mathcal{Q}({o},\cdot))\right] 
\end{equation}
where $\mathcal{Q}({o},\cdot)\propto$ exp $\{\frac{1}{\alpha} Q({o},\cdot)\}.$
The parameters of critic network is updated through the loss function $J(Q)$, which can be expressed as 
\begin{equation}    J(Q)=\mathbb{E}_{({o},{a},r,{o}^{'})\thicksim\mathcal{B}}\left[(Q({o},{a})-r-\gamma\bar{V}({o}^{'}))^{2}\right] 
\end{equation}
We update the encoder $p_{\phi}$ and decoder $g_{\varphi}$ with the following objective:
\begin{equation}
	 J(RAE)=\mathbb{E}_{{x}}\left[\log p_{\phi}({x}|{z})+\lambda_{{z}}\|{z}\|^{2}+\lambda_{\phi}\|\phi\|^{2}\right]
\end{equation}

\subsection{Causal Feature Selection}

\begin{figure}
    \begin{center}
        \includegraphics[width=1\linewidth]{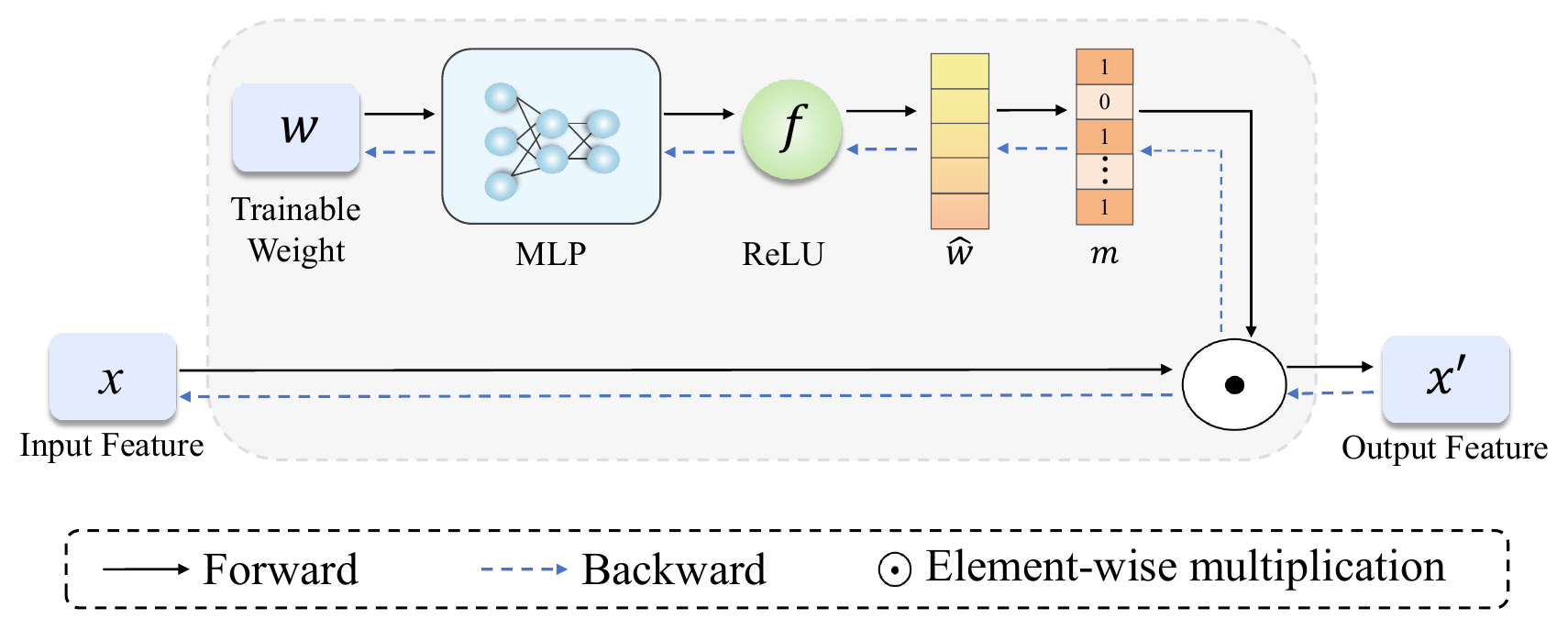}
    \end{center}
    \caption{\textbf{Causal Feature Selection Module}. This module transforms a trainable weight into a binary mask for feature selection.}
    \label{fig:cfs}
    \vspace{-4mm}
\end{figure}

\begin{figure*}[t]
    \centering
    \includegraphics[width=0.8\linewidth]{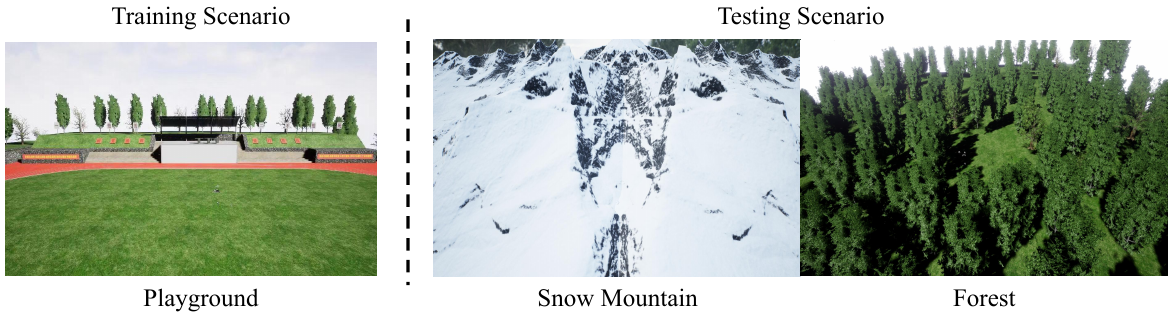}
    \vspace{-2mm}
    \caption{\textbf{Simulation scenarios for model training and testing}. Specifically, playground scenario is used for model training, while grassland, snow mountain and forest scenarios are used for testing.}
    \label{fig:scenarios}
    \vspace{-3mm}
\end{figure*}

To filter out non-causal representation factors and select causal components, we design a plug-and-play and lightweight causal feature selection (CFS) module, which can be embedded into the actor network.
The CFS module generates a differentiable binary mask for channel selection, which can explicitly suppress the influence of non-causal channels.
As shown in Fig.~\ref{fig:cfs}, to perform causal feature selection, we multiple the input feature $x \in \mathbb{R}^{C}$ with the generated binary mask $m$ as follows:
\begin{equation}
	x^{'}=x \odot m,
\end{equation}
where $C$ is the number of feature channels. In this way, the binary mask $m$ can explicitly activate the causal feature channels and deactivate others. After that, the modulated feature $x^{'}$ can eliminates the effects of non-causal representation factors and improve generalization ability.

The essence of the CFS module lies in generating a differentiable binary mask $m$, which can be integrated into the actor network and trained end-to-end.
Drawing inspiration from prior successful implementations~\cite{you2019gate,ramakrishnan2020differentiable}, we devise a differentiable mask generation process. Given an intermediate vector $x \in \mathbb{R}^{C}$ within the actor network as input, we assign a trainable weight $w \in \mathbb{R}^{C}$ and the corresponding mask $m$ is determined as follows:
\begin{equation}
    \hat{w} = ReLU(MLP(w)),
\end{equation}
\begin{equation}
    m=\frac{\hat{w}^{2}}{\hat{w}^{2}+\epsilon},
\end{equation}
where $\epsilon$ is an infinitesimally small positive number.
We first generate an intermediate variant $\hat{w}$ by transforming the input weight $w$ with a small MLP and then applying a ReLU activation.
For each channel, it is evident that $m$ contains a value of $0$ if $\hat{w}$ is $0$, and otherwise, it contains $1$, given that $\epsilon$ is an extremely small number.
Following this, the function transforms the trainable weight $w$ into a differentiable binary mask $m$ without the need for additional manual threshold design.

\begin{figure*}[t]
    \begin{center}
        \includegraphics[width=0.81\linewidth]{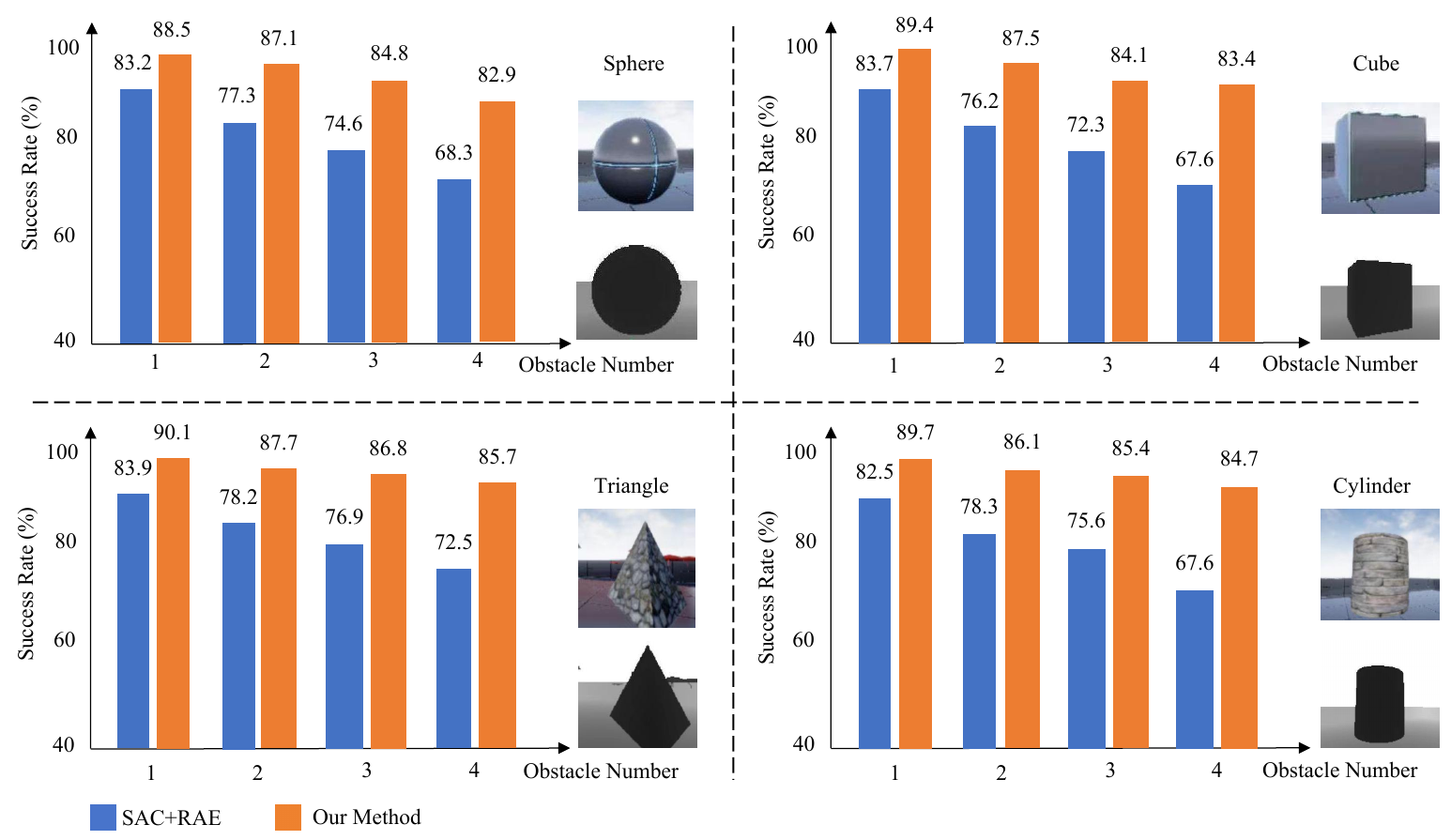}
    \end{center}
    \vspace{-2mm}
    \caption{\textbf{Performance comparison with different unseen obstacles}. Our method can achieve consistent performance improvement over SAC+RAE.}
    \label{fig:obstacle}
    \vspace{-2mm}
\end{figure*}

To ensure that the CFS module can learn a binary mask to filter out non-causal feature channels while still maintaining the ability to achieve an effective collision avoidance strategy, we use $J(\pi)$ to update the trainable weights in the CFS module when calculating binary masks, \ie, the pre-set weights and the small MLP.


\input{table/background}

\input{table/background_obstacle}

\section{EXPERIMENT AND RESULTS}

\subsection{Simulation Environment and Experiments Setup}
We create multiple simulation environments in Unreal Engine~\cite{qiu2016unrealcv} and conduct simulated experiments using the AirSim~\cite{madaan2020airsim} simulator.
All experiments are conducted on a device equipped with Ubuntu 20.04 operating system, intel i9-12900k CPU, and a single NVIDIA RTX 3090 GPU.
We construct a policy learning model based on PyTorch and employ ROS~\cite{quigley2009ros} to acquire sensor data and control the motion of the UAVs.

\subsection{Performance Metrics and Experiment Scenarios}
\subsubsection{Performance Metrics}
Building upon the previous work~\cite{huang2022vision}, we employ the following performance indicators for evaluation:
\begin{itemize}
\item Success Rate:
The percentage of agent that successfully reach their corresponding target position without collisions within a specified number of time steps. 
\item SPL (Success weighted by Path Length):
The proportion of UAVs that successfully reach the target positions, considering the path length, calculated across the test scenario with $N$ UAVs and $M$ episodes as follows: 
\begin{equation}
    SPL=\frac{1}{N} \frac{1}{M} \sum_{i = 1}^{N}\sum_{j = 1}^{M}S_{i,j}\frac{l_{i,j}}{max(p_{i,j}, l_{i,j})}   
\end{equation}
 where $l_{i,j}$ denotes the shortest-path distance from the $i^{th}$ UAV's initial position to the target position in episode $j$, and $p_{i,j}$ represents the actual path length traversed by the $i^{th}$ UAV in this episode. Additionally,  $S_{i,j}$ is a binary indicator denoting success or failure in this episode.
\item Extra Distance:
The average additional distance traveled by UAVs relative to the straight-line distance between the initial and goal positions.
\item Average Speed:
The average speed of all UAVs during testing, obtained from the average speed of each episode. 
\end{itemize}

\subsubsection{Scenarios}
To investigate the generalization ability of DRL-based multi-UAV collision avoidance systems, we design several typical testing scenarios by changing backgrounds or adding unseen obstacles, which differ from the training scenario.
Specifically, as illustrated in Fig.~\ref{fig:scenarios}, we train the model in a playground scenario without any obstacles.
To evaluate the generalization ability with respect to unseen backgrounds, we design two scenarios, \ie, snow mountain and forest, inspired by practical UAV applications such as wildlife monitoring~\cite{gonzalez2016unmanned}, reconnaissance and surveillance~\cite{zhang2020method}.
To assess the generalization ability concerning unseen obstacles, we design four types of obstacles with different common shapes, \ie, cube, sphere, triangle and cylinder.





\subsubsection{Initialization}
\begin{itemize}
\item Random Pattern: The flight range of each UAV is restricted within a space of dimensions $(16*16*4)$. The initial and target positions of the UAVs are randomly generated within this range.
\item Cycle Pattern:
The initial positions of each UAV are uniformly distributed within a circular area at the same height. The target positions of each UAV are set on the opposite side of the circular area from their initial positions.
\end{itemize}

\subsection{Performance Comparison}
To clearly evaluate the generalization ability of our method, we conduct performance comparison with the SOTA method, \ie, SAC+RAE, under different unseen testing scenarios by adding different backgrounds and obstacles.

\subsubsection{Unseen Background}

As shown in TABLE~\ref{tab:background}, our method consistently improves the navigation success rate and SPL, and achieves a higher average speed across both seen and unseen backgrounds. This demonstrates the effectiveness of our method in enhancing generalization ability.
It is notable that the flight path planned by our method is slightly longer than of SAC+RAE, because the planed path includes more collision avoidance actions.

\subsubsection{Unseen Obstacle}

To evaluate the adaptability to unseen obstacles, we add typical obstacles of various shapes and quantities in the playground background.
As shown in Fig.~\ref{fig:obstacle}, our method consistently outperforms SAC+RAE, when encountering different obstacles.

\subsubsection{Unseen Background and Obstacle}

To further evaluate the generalization ability, we construct more challenging scenarios by combining unseen backgrounds with four cylinder obstacles.
As shown in TABLE~\ref{tab:background_obstacle}, our method achieves a significant improvement over SAC+RAE.

\subsection{Ablation Study}
In this subsection, we conduct experiments to demonstrate the effectiveness of our proposed module. All experiments are specifically carried out in the forest scenario with random initialization by default.

\subsubsection{Casual Feature Selection}

\begin{figure*}
    \centering
    \includegraphics[width=0.8\linewidth]{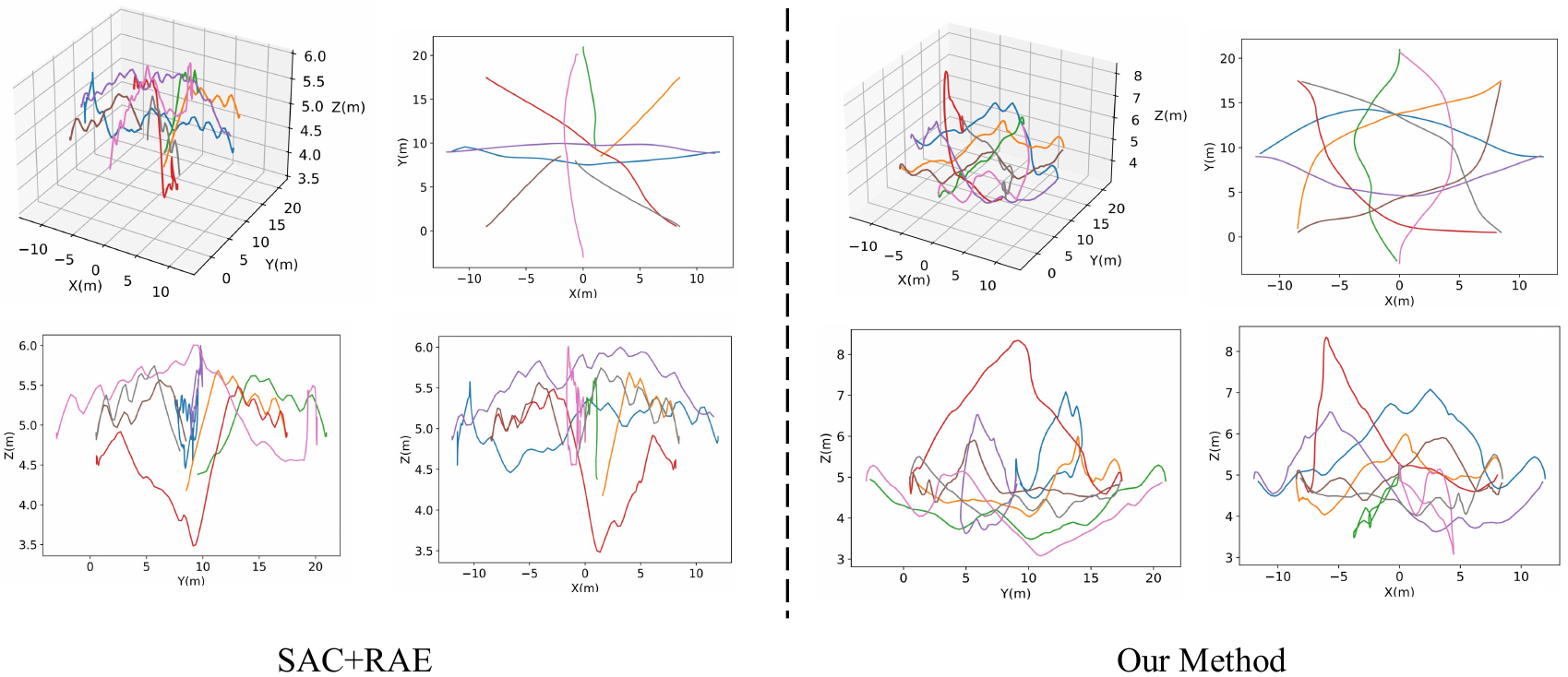}
    \caption{\textbf{Visualization of UAV trajectories in perspective drawing and three-view drawing}. The trajectories of different UAVs are represented by different colors. Best viewed in color.}
    \label{fig:trajectory}
\end{figure*}

\begin{figure}
    \begin{center}
        \includegraphics[width=0.85\linewidth]{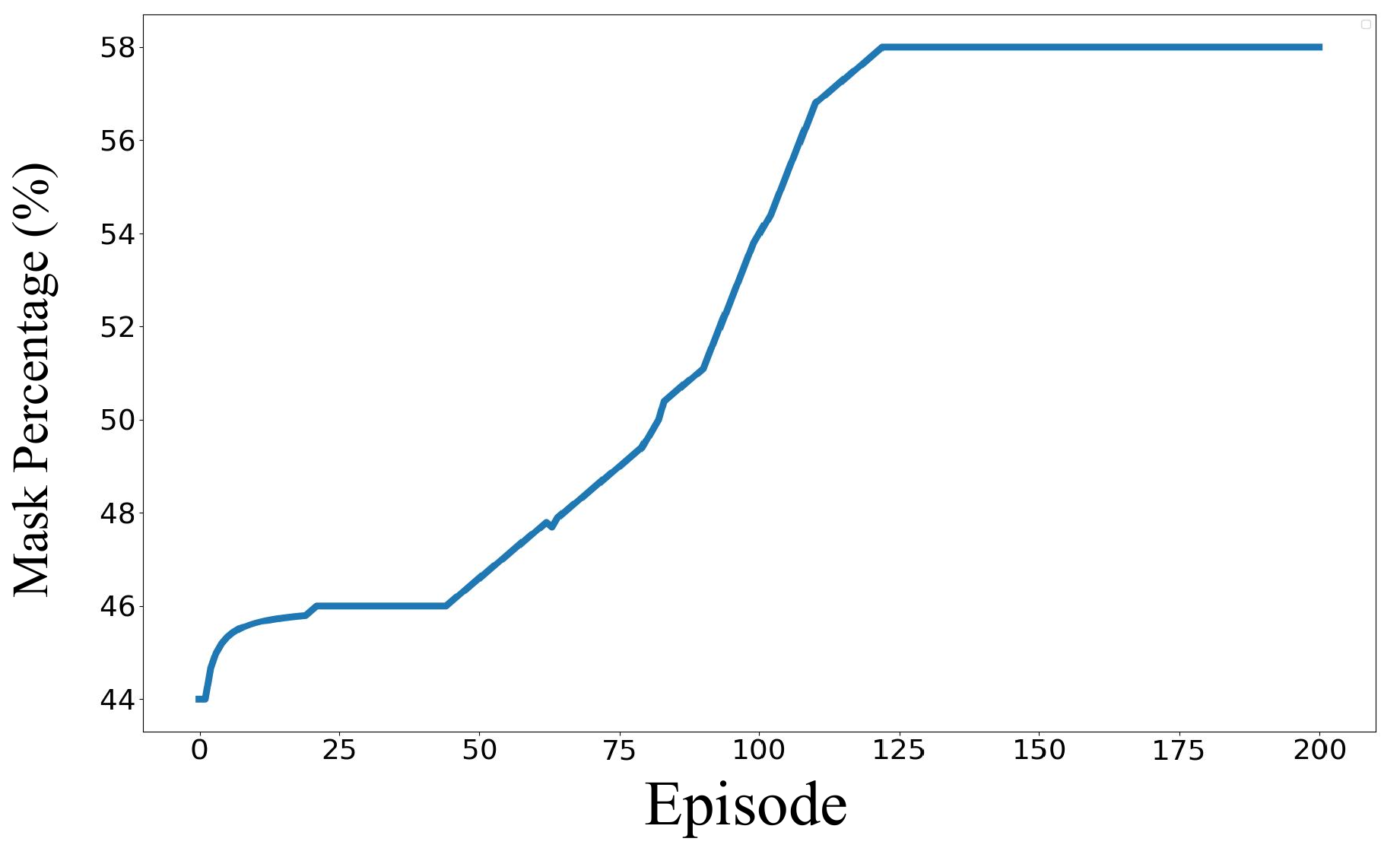}
    \end{center}
    \caption{\textbf{Visualization on variation of generated binary masks in CFS module}. During training, the percentage of zero value in the binary mask consistently increases and finally converges.}
    \label{fig:mask}
    \vspace{-4mm}
\end{figure}

\input{table/circle_initialization}
\input{table/number_scalability}

The key to our method's improvement in generalization ability lies in the causal feature selection mechanism, which employs a differentiable binary mask.
To clearly illustrate the feature selection process, we take the first CFS module as an example and visualize the percentage of zero values in the generated binary mask during training, as shown in Fig.~\ref{fig:mask}.
Throughout the training, the percentage consistently increases and eventually stabilizes at around 58\%, indicating that the CFS module can effectively identify and eliminate non-causal feature channels, ultimately filtering out near half of the feature channels.

\subsubsection{Circle Initialization}
As demonstrated in TABLE~\ref{tab:circle_initialization}, we evaluate the collision avoidance navigation performance of UAVs using circle initialization, a commonly used assessment method. Our method significantly outperforms SAC+RAE, achieving a much higher success rate.

\subsubsection{Scalability}

To assess the scalability of our method, we configure scenarios with 8, 10, 12, and 14 UAVs in an
unseen forest environment. 
As demonstrated in TABLE~\ref{tab:scale}, our method consistently outperforms SAC+RAE, showcasing its
superior robustness and scalability.

\subsubsection{Trajectory Visualization}
To validate the quality of the planed path, we provide visualization of UAV trajectories. As shown in Fig.~\ref{fig:trajectory}, our method can achieve smoother and more complete flight trajectories, while SAC+RAE results in collision of 4 UAVs in unseen scenarios. 
The results demonstrate that our method can generate a more robust and effective flight strategy.

\section{CONCLUSIONS}

In this paper, we propose a robust policy learning approach by designing a causal feature selection (CFS) module to enhance the generalization ability of DRL techniques in unseen scenarios. 
The CFS module can be integrated into the policy network and effectively filters out non-causal components during representation learning, thereby reducing the influence of spurious correlations. 
To demonstrate the generalization ability, we conduct extensive experiments in various testing scenarios by altering backgrounds and introducing unseen obstacles. 
The experimental results indicate that our method significantly outperforms the previous SOTA method, thereby showcasing excellent generalization ability and robustness.










\bibliographystyle{IEEEtran}
\small\bibliography{reference}

\end{document}

%% file: table/background.tex
\begin{table*}[t]
    \renewcommand{\arraystretch}{1.1}
    \caption{Performance (as mean/std) comparison under different backgrounds.}
    \vspace{-3mm}
    \label{tab:background}
    \begin{center}
        \begin{tabular}{ccccccc}
            \toprule
            Scene & Seen/Unseen & Method & Success Rate (\%) & SPL (\%) & Extra Distance (m) & Average Speed (m/s)\\
            \midrule
            \multirow{2}[1]{*}{Playground} &\multirow{2}[1]{*}{Seen} & SAC+RAE & 90.6 & 78.5 & \textbf{1.262/1.242} & 0.890/0.185\\
            &  & Our method & \textbf{94.1} ($\uparrow 3.5$)  & \textbf{80.3} ($\uparrow 1.8$) & 1.662/1.389 & \textbf{0.983/0.137} \\
            \hline
            \multirow{2}[1]{*}{Snow Mountain} &\multirow{2}[1]{*}{Unseen} & SAC+RAE & 82.1 & 71.4 & \textbf{1.658/1.682} & 0.852/0.186 \\
            &  & Our method & \textbf{90.6} ($\uparrow 8.5$)  & \textbf{77.4} ($\uparrow 6.0$) & 1.749/1.568 & \textbf{0.919/0.142} \\
            \hline
            \multirow{2}[1]{*}{Forest} &\multirow{2}[1]{*}{Unseen} & SAC+RAE & 69.1 & 58.6 & \textbf{1.746/1.653} & 0.803/0.145 \\
            &  & Our method & \textbf{86.8} ($\uparrow 17.7$)  & \textbf{72.5} ($\uparrow 13.9$) & 2.083/2.335 & \textbf{0.842/0.181}\\
            
            \bottomrule
        \end{tabular}              
    \end{center}
    \vspace{-2mm}
 \end{table*}

%% file: table/background_obstacle.tex
\begin{table*}[t]
    \renewcommand{\arraystretch}{1.1}
    \caption{Performance comparison under different unseen backgrounds and with 4 cylinder obstacles.}
    \vspace{-3mm}
    \label{tab:background_obstacle}
    \begin{center}
        \begin{tabular}{ccccccc}
            \toprule
            Scene & Seen/Unseen & Method & Success Rate (\%) & SPL (\%) & Extra Distance (m) & Average Speed (m/s)\\
            \midrule
            \multirow{2}[1]{*}{Snow Mountain} &\multirow{2}[1]{*}{Unseen} & SAC+RAE & 61.3 & 50.6 & \textbf{2.093/1.982} & 0.787/0.208 \\
            &  & Our method & \textbf{81.1} ($\uparrow 19.8$)  & \textbf{64.0} ($\uparrow 13.4$) & 2.694/2.503 & \textbf{0.839/0.186} \\
            \hline
            \multirow{2}[1]{*}{Forest} &\multirow{2}[1]{*}{Unseen} & SAC+RAE & 50.6 & 41.9 & \textbf{2.314/2.026} & 0.682/0.106 \\
            &  & Our method & \textbf{76.5} ($\uparrow 25.9$)  & \textbf{63.8} ($\uparrow 21.9$) & 2.474/2.365 & \textbf{0.718/0.122} \\
            \bottomrule
        \end{tabular}              
    \end{center}
    \vspace{-2mm}
 \end{table*}

%% file: table/circle_initialization.tex
 \begin{table*}
    \caption{Performance comparison under circle initialization.}
    \vspace{-3mm}
    \label{tab:circle_initialization}
    \begin{center}
        \begin{tabular}{cccccc}
            \toprule
            Scene & Method & Success Rate (\%) & SPL (\%) & Extra Distance (m) & Average Speed (m/s)\\
            \midrule
            \multirow{2}[1]{*}{Forest} & SAC+RAE & 60.2 & 59.4 & \textbf{0.733/0.925} & \textbf{1.164/0.132}\\
            & Our method & \textbf{81.5} ($\uparrow 21.3$)  & \textbf{72.7} ($\uparrow 13.3$) & 3.076/1.208 & 1.109/0.087 \\
            \bottomrule
            \end{tabular}                                                                                                                                  
    \end{center}
 \end{table*}

%% file: table/number_scalability.tex
\begin{table*}[t]
    \renewcommand{\arraystretch}{1.1}
    \vspace{-3mm}
    \caption{Performance comparison with different numbers of UAVs.}
    \vspace{-3mm}
    \label{tab:scale}
    \begin{center}
        \begin{tabular}{cccccc}
            \toprule
            Number& Method & Success Rate (\%) & SPL (\%) & Extra Distance (m) & Average Speed (m/s)\\
            \midrule
            \multirow{2}[1]{*}{8} & SAC+RAE & 60.8 & 49.1 & \textbf{2.305/2.133} & 0.739/0.203 \\ 
            & Our method & \textbf{79.2} ($\uparrow 18.4$)  & \textbf{63.7} ($\uparrow 14.6$) & 2.707/2.641 & \textbf{0.828/0.184}\\
            \hline
            \multirow{2}[1]{*}{10} & SAC+RAE & 59.2 & 48.3 & \textbf{2.518/2.271} & 0.650/0.175 \\
            & Our method & \textbf{77.5} ($\uparrow 18.3$)  & \textbf{60.9} ($\uparrow 12.6$) & 2.746/2.456 & \textbf{0.733/0.149} \\
            \hline
            \multirow{2}[1]{*}{12} & SAC+RAE & 56.2 & 45.8 & \textbf{2.626/2.377} & 0.587/0.161 \\
            & Our method & \textbf{76.4} ($\uparrow 20.2$)  & \textbf{57.9} ($\uparrow 12.1$) & 3.238/2.901 & \textbf{0.655/0.136} \\
            \hline
            \multirow{2}[1]{*}{14} & SAC+RAE & 53.6 & 42.1 & \textbf{2.950/2.662} & 0.508/0.142 \\ 
            & Our method & \textbf{73.4} ($\uparrow 19.8$)  & \textbf{55.3} ($\uparrow 13.2$) & 3.249/2.787 & \textbf{0.580/0.119} \\
            \bottomrule
            \end{tabular}
    \end{center}
    \vspace{-4mm}
 \end{table*}